\def\year{2018}\relax
\documentclass[letterpaper]{article} 
\usepackage{aaai18}  
\usepackage{times}  
\usepackage{helvet}  
\usepackage{courier}  
\usepackage{url}  
\usepackage{graphicx}  
\usepackage{amsmath, amssymb}
\usepackage[utf8]{inputenc} 
\usepackage[T1]{fontenc}    
\usepackage{hyperref}       
\usepackage{amsfonts}       
\begin{document}
%
\title{RRA: Recurrent Residual Attention for Sequence Learning}
\author{Cheng Wang\\
NEC Laboratories Europe, Heidelberg, Germany\\
cheng.wang@neclab.eu\\
}
\maketitle
\begin{abstract}
In this paper, we propose a recurrent neural network (RNN) with residual attention (RRA) to learn long-range dependencies from sequential data.
We propose to add residual connections across timesteps to RNN, which explicitly enhances the interaction between current state and hidden states that are several timesteps apart. This also allows training errors to be directly back-propagated through residual connections and effectively alleviates the gradient vanishing problem.  
We further reformulate an attention mechanism over residual connections. An attention gate is defined to summarize the individual contribution from multiple previous hidden states in computing the current state. 
We evaluate RRA on three tasks: the adding problem, pixel-by-pixel MNIST classification and sentiment analysis on the IMDB dataset.  
Our experiments demonstrate that RRA yields better performance, faster convergence and more stable training compared to a standard LSTM network. Furthermore, RRA shows highly competitive performance to the state-of-the-art methods.
\end{abstract}

\

\section{Introduction}
Deep neural networks (DNN)  have shown significant improvements in several application domains including image recognition \cite{krizhevsky2012imagenet}, natural language processing \cite{mikolov2013distributed} and speech recognition \cite{hinton2012deep}. Recurrent neural networks (RNNs), a particular type of DNN, have powerful capability in processing complicated sequential data. By using recurrent connections, the previous context information can be captured and used to predict the next hidden state output.  However, training RNN remains a difficult task due to gradient vanishing and exploding problems \cite{pascanu2013difficulty}, especially when the RNN needs to learn very long dependencies from sequential inputs. 
The main issue is that training an RNN using back-propagation through time (BPTT) \cite{williams1986learning} entails multiplying gradients a large number of times (specifically, once for each time step) with the weights matrix $\mathbf{W}$. 
If $\mathbf{W}$ contains small values (namely, if the largest eigenvalue of $\mathbf{W}$ is less than 1), then gradient contributions from “far away” states become zero and have no influence on future states, this is the \textit{gradient vanishing problem}. On the other hand, if the weights in the matrix are large, the gradient signal grows without bound, and learning diverges, this is the \textit{gradient exploding problem}.  To alleviate the effects of gradient vanishing, many methods have been proposed. Long Short-Term Memory (LSTM) \cite{hochreiter1997long} can be seen as the most successful one among those techniques.    
The introduced memory cell in LSTM has its own input, forget and output gates to control whether to store the context information or remove it from memory. This allows LSTM networks to capture the long-range relational dependencies from input sequences as compared to a regular RNN. 

\begin{figure}
\centering
\includegraphics[width=0.5\textwidth]{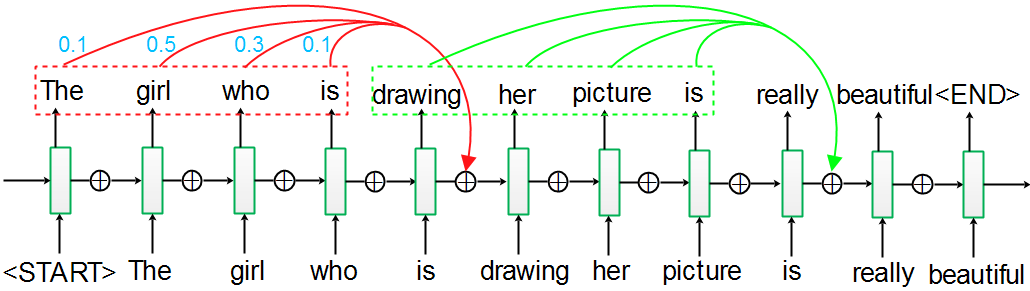}
\caption{Learning recurrent residual attention. The interaction with hidden states that far apart can be enhanced by residual connections. The attention over residual connections decides how far RRA cell can look back at given timestep, meanwhile, controls the individual contribution of previous hidden states. In this example, each RRA cell is able to look back at the past 5 time steps, the semantic dependency between the word ``girl'' and ``her'' can be explicitly captured.}
\label{fig:RRA_framework}
\end{figure} 

The gradient vanishing problem is not limited to recurrent neural network and can also appear in feedforward neural network, particularly, in training very deep networks.  If we treat an RNN in its unfolded form, a shallow RNN with multiple timesteps is equivalent to a very deep network. Residual learning \cite{he2016deep} provides a novel learning scheme for ultra-deep convolutional neural network (CNN) (e.g. more than 1000 layers) by introducing residual connections across layers.  These \emph{shortcut connections} connect far-away layers to ensure training error signal can be back-propagated from higher layer to lower layer directly and alleviate gradient vanishing problem. Inspired by the success of residual learning in CNN on computer vision tasks, this work reformulates residual learning into recurrent network for learning ultra-long range dependencies across timesteps in sequence learning.

Different to residual learning \cite{he2016deep} where an identity shortcut connection is used to add the input and the outputs from stacked layers (i.e. $\mathcal{F}(\mathbf{x})$+$\mathbf{x}$, $\mathcal{F}$ is residual function), in the context of sequence learning,  we reformulate the recurrent residual connection to have attention over multiple precessing steps. It results in a residual function with attention across timesteps: $\mathcal{M}(\mathbf{x}_t,\mathbf{h}_{t-1})$+$\mathcal{F}(\mathbf{h}_{t-2}, \mathbf{h}_{t-3},..., \mathbf{h}_{t-K-1};\mathbf{W}_a)$ where $\mathcal{M}$ is a recurrent model and $\mathbf{W}_a$ is the attention weights. At each timestep $t$, in computing the current state $\mathbf{h}_t$, this reformulation ensures recurrent units have the ability to look back as far as $K$+$1$ past timesteps and control the relative contribution of each hidden state $\mathbf{h}_{t-2}, \mathbf{h}_{t-3},..., \mathbf{h}_{t-K-1}$ to the current state $\mathbf{h}_t$. 

Even though attention mechanism has been widely studied in machine translation \cite{bahdanau2014neural}, image captioning \cite{xu2015show}, object detection \cite{ba2014multiple} and generative models \cite{mnih2014recurrent,gregor2015draw}. Basically, this sort of attention models are either layer-based or network-based. They are only allowed to receive attended information from a previous layer or a separate network.
By casting attention mechanism to recurrent residual connection, the recurrent unit provides a more natural way to sequence learning. Because it explicitly looks back at multiple preceding steps and automatically decides how much previous information should be ``seen'' by weighting them. For a specific sequential pattern (e.g. English or German sentence $w_1,...w_{T}$), the semantic dependencies between words that are far apart (e.g. $w_{t}$ and $w_{t-k}$, 1$<$$k$$<$$t$) can be stronger than that between two adjacent words (e.g. $w_t$ and $w_{t-1}$). Figure \ref{fig:RRA_framework} gives an example which intuitively supports our assumption.  The word ``\emph{drawing}'' is explicitly involved in predicting the word ``\emph{her}'',   it is obvious that word ``\emph{girl}'' would also make significant semantic contribution. Essentially, the sentence is saying: ``\emph{The girl is beautiful}'', however, regular RNNs suffer difficulties in capturing the meaning. Thus, it is reasonable to explicitly consider the information that are several steps apart in learning the semantic meaning from sequential data. In this work, we address this problem by casting attention mechanism to residual connection over timesteps in recurrent network. 

The benefits of recurrent residual attention (RRA) are two fold: (1) RRA enhances the interactions between hidden states that are several steps apart, that is, RRA allows training error can be back-propagated across multiple timesteps. (2) The attention over residual connection gives a more natural way in which past hidden states can selectively ``attend'' to future states in sequence learning. 

Our main contributions are summarized as follows:  

\begin{itemize}
\item We propose a novel learning scheme for sequential data, it reformulates residual learning with attention in recurrent network. The code will be made publicly available soon.
\item A new gate---\emph{attention gate} is defined in LSTM RNN to control the individual contribution of context information from multiple previous hidden states.
\item Our proposed RRA shows promising performance as compared to a standard LSTM network on three benchmark tasks: the adding problem, pixel-by-pixel MNIST and sentiment analysis. RRA also outperforms or matches the state-of-the-art methods.
\end{itemize}

The rest of this paper is structured as follows, section 2\ref{sec:relate_work} gives the related work. In section 3\ref{sec:models}, we elaborate the reformation of residual learning with attention in recurrent manner. We describe our experiments and discussions in section 4\ref{sec:experiment} and conclude this work in section 5\ref{sec:conclusion}.

\section{Related Work}
\label{sec:relate_work}
\textbf{Recurrent Neural Network (RNN)}~~ RNN is a powerful network architecture for processing sequential data. 
It has been widely used in natural language processing \cite{socher2011parsing}, speech recognition \cite{graves2013speech} and handwriting recognition \cite{graves2009novel} in recent years. 
In RNN, it allows cyclical connection and reuse the weights across different instances of neurons, each of them associated with different time steps.
This idea can explicitly support network to learn the entire history of previous states and map them to current states. 
With this property, RNN is able to map an arbitrary length sequence to a fixed length vector. 
But RNN is known for its difficult training due to gradient vanishing problem.

The vanishing problem was originally found in \cite{hochreiter1997long}, then LSTM (Long short-term memory) was proposed to prevent gradient from vanishing during training. Therefore, compare to traditional RNN, LSTM has the ability to learn the long-term dependencies between inputs and outputs. 
Recently, LSTM has became very popular in the field of machine translation \cite{cho2014learning}, speech recognition \cite{graves2013speech} and sequence learning \cite{sutskever2014sequence} recently. Another special type of RNN is Gated Recurrent Unit (GRU)\cite{cho2014learning}. It simplifies LSTM by removing memory cell and provides a different way to prevent vanishing gradient problem. Our work falls into this category and aims to alleviate gradient vanishing in learning ultra-long dependencies.
\newline

\noindent  \textbf{Residual Learning}
Previous work \cite{simonyan2014very,szegedy2015going} have proven that network depth is of crucial importance of neural network architectures, but it is more challenging to train deeper networks. Residual learning \cite{he2016deep}  paves a way for training such networks.  The residual mapping between layers enables networks can be substantially deep (e.g. with hundreds of layers) and leads more efficient optimization, most importantly, yields better performance. The short-cut skip connections were considered across multiple layers to force a direct information flow in both forward and backward passes.  By doing this, feedforward signals as well as feedback errors can be passed easily. Adding residual connection across layers has shown its powerful capability in computer vision \cite{he2016deep,szegedy2017inception}. Inspired by this, our work incorporates residual connection across multiple precessing steps to learn long and complex dependencies from sequential data.
\begin{figure}
\centering
\includegraphics[width=0.5\textwidth]{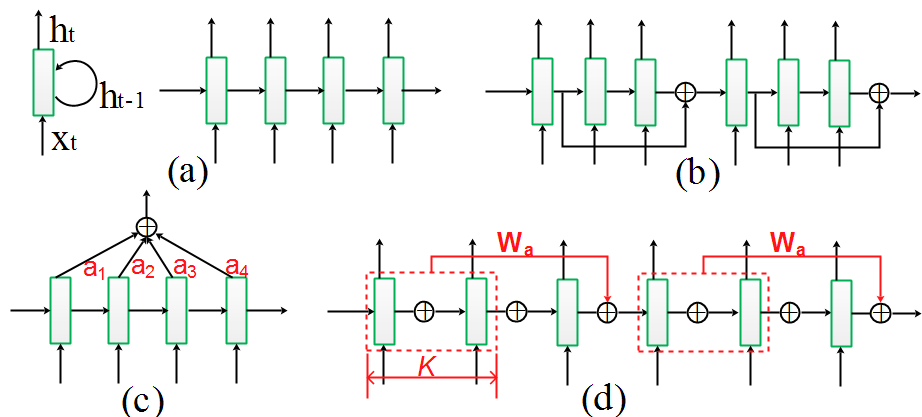}
\caption{Overview of proposed methods. (a) Standard RNN and its unfolded form. (b) RNN with residual connections. (c) recurrent network with attention mechanism (over layers). (d) Recurrent residual with attention (over timesteps), at each timestep $t$, units are able to look back at the past $K$+$1$ states in computing the current state $\mathbf{h}_t$,  and $\sum_{k}^{K}\mathbf{W}a_k=1$. }
\label{fig:framework}
\end{figure} 
\newline

\noindent \textbf{Attention Mechanism}
Attention in neural networks \cite{bahdanau2014neural} is designed to assign weights to different inputs instead of threat all input sequences equally as original neural networks do. It can be seen as an additional network that is now widely incorporated into different neural networks leading to a new variety of models \cite{xu2015show,ba2014multiple,mnih2014recurrent,gregor2015draw}.  Formally, an attention model takes $k$ arguments e.g. $h_1$,...,$h_k$, and a context information $c$. It returns a weighted output $z$ which summaries based on how $h_i$ is related to context $c$. The weights are corresponds to the relevances between each $h_i$ and $c$ and sum to 1, e.g. the weights $a_k$ in Figure \ref{fig:framework} (c).  This determines the relative contributions of each $h_i$ to final output.  But the current state-of-the-art attention methods are either layer or network based, and not well studied in recurrent manner. This work reformulates an attention over residual connection in recurrent network.
\vspace{-0.2cm}
\section{Models}
\label{sec:models}
This section describes our proposed approach to learn recurrent residual attention from sequential data. 
We firstly introduce existing way for sequence learning with recurrent network and explain our intuition of extending recurrent network to learn more complex dependencies. Then we describe how to reformulate residual connection into RNN, and followed by casting attention mechanism to recurrent residual connection. Here, we use LSTM as base recurrent network to elaborate our approach, but it can be easily generalized to plain RNN or GRU.

\subsection{Recurrent Networks for Sequence Learning}
A recurrent network basically generalizes feedforward network to learning from sequential data. The goal of recurrent models is to estimate the conditional probability $p(\mathbf{y}_1,...\mathbf{y}_{T'}|\mathbf{x}_1,..\mathbf{x}_T)$ by:
\begin{align}
p(\mathbf{y}_1,...\mathbf{y}_{T'}|\mathbf{x}_1,...\mathbf{x}_T)=\prod\nolimits_{t=1}^{T'}p(\mathbf{y}_t|\mathbf{y}_1,...,\mathbf{y}_{t-1})\\
p(\mathbf{y}_t|\mathbf{y}_1,...,\mathbf{y}_{t-1})=p(\mathbf{y}_t|\mathbf{h}_t)\\
\mathbf{h}_t=\mathcal{M}(\mathbf{h}_{t-1},\mathbf{x}_t)
\label{equ:recu}
\end{align}
where $(\mathbf{x}_1,...\mathbf{x}_T)$ and $(\mathbf{y}_1,...\mathbf{y}_{T'})$ are input sequence and target sequence respectively. The input sequence length $T$ may differ from target sequence length $T'$. $\mathbf{h}_t$ is the hidden state from a model $\mathcal{M}$ for a given hidden state $\mathbf{h}_{t-1}$ and a new input $\mathbf{x}_t$. The $\mathcal{M}$ is recurrent model that can be a standard RNN or its variants. The equation (\ref{equ:recu}) can be viewed as a general form of recurrent learning algorithm which is able to capture the semantic dependencies across timesteps. For example the hidden state $\mathbf{h}_{t-1}$ is explicitly used for outputting $\mathbf{h}_{t}$ while the past hidden state before $\mathbf{h}_{t-1}$ are only implicitly involved. 

This challenges existing RNNs in a task that needs model to explicitly capture the long-range semantic dependencies between the states that are several timesteps apart,
as the task we described in Figure \ref{fig:RRA_framework}. 
Adding a shortcut connection to skip one or multiple timesteps and enforcing a direct information across timesteps is a way to explicitly use previous hidden states in ($\mathbf{h}_{2}$,...,$\mathbf{h}_{t-k-1}$) in computing future states. This entails recurrent residual learning.
 
\subsection{Recurrent Residual Learning}
The overview of reformulating recurrent network to have residual connection is illustrated in Figure \ref{fig:framework} (b), in which a shortcut connection is designed to impose a fluent information flow across timesteps. With residual connection in recurrent network, at a given timestep $t$, the hidden state $\mathbf{h}_t$ can be computed as:
\begin{equation}
\mathbf{h}_t=\mathcal{M}(\mathbf{h}_{t-1},\mathbf{x}_{t};\mathbf{W}_m)+\mathcal{F}(\mathbf{h}_{t-k};\mathbf{W}_f)
\label{equ:1}
\end{equation}
where $\mathcal{M}$ is a  RNN model with weights $\mathbf{W}_m$, it receives $\mathbf{h}_{t-1}$ and $\mathbf{x}_{t}$ as regular RNN. Here we keep $\mathcal{M}$ to receive $\mathbf{h}_{t-1}$ so as to form a residual skip connection across timesteps.  $\mathcal{F}$ approximates
the residual function with weights $\mathbf{W}_f$. $\mathcal{F}$ can be an identity function such that $\mathcal{F}(\mathbf{h}_{t-k};\mathbf{W}_f)$ = $\mathbf{h}_{t-k}$ where $\mathbf{h}_{t-k}$ is the hidden state at $t$-$k$ time step. With this formulation, when computing a hidden state $\mathbf{h}_{t}$,  besides $\mathbf{h}_{t-1}$ and $x$, $\mathbf{h}_{t-k}$ can be explicitly considered. If  $\mathbf{W}_f$ approximating 0, equation (\ref{equ:1}) returns back to plain RNN. 

By making $\mathcal{F}$ to weight multiple previous hidden states, i.e. $\mathbf{h}_{t-2}$,...,$\mathbf{h}_{t-k}$, can lead to recurrent residual learning with attention over timesteps:
\begin{equation}
\mathbf{h}_t=\mathcal{M}(\mathbf{h}_{t-1},\mathbf{x}_{t};\mathbf{W}_m)+\mathcal{F}(\mathbf{h}_{t-2},...,\mathbf{h}_{t-k};\mathbf{W}_a)
\label{equ:2}
\end{equation}
where $\mathbf{W}_a$$\in$$\mathbb{R}^{1 \times (k-1)}$ is the attention weight matrix that controls the relative contribution of the past hidden states and $\sum_{i=1}^{k-1}\mathbf{W}_a^{(i)}$=$1$.
\begin{figure}
\centering
\includegraphics[width=0.4\textwidth]{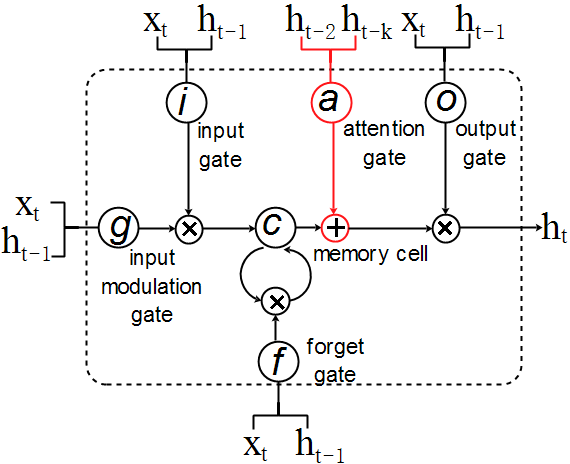}
\caption{RRA cell.  An attention gate is defined to control how much information from hidden state $\mathbf{h}_{t-2}$ to $\mathbf{h}_{t-k}$ should be considered in computing current state $\mathbf{h}_t$. }
\label{fig:RRA_cell}
\end{figure} 

\subsection{Learning Recurrent Residual Attention}
Figure \ref{fig:framework} (d) gives our design of reformulating attention on residual connections in recurrent network. The recurrent residual attention is considered at each timestep, this can be viewed as a sliding attention window with size of $K$ over timesteps. 
To make the past states selectively ``attend'' in future state, we enforce the residual attention effects memory cell directly,  a new gate---\emph{attention} gate is defined to LSTM cell, making LSTM has residual attention. Then the equation (\ref{equ:2}) is reformulated as
\begin{equation}
\mathbf{h}_t=\mathcal{M}((\mathbf{h}_{t-1},\mathbf{x}_{t}, \mathbf{a}_t);\mathbf{W}_m)
\label{equ:3}
\end{equation}
where $\mathbf{a}_t$=$\mathcal{F}(\mathbf{h}_{t-2},...,\mathbf{h}_{t-k};\mathbf{W}_a)$.
 Figure \ref{fig:RRA_cell} demonstrates the internal gates of RRA cell, where the attention gate controls the relative contributions of  the past $K$ states. Basically, the hidden state of each gate within RRA can be computed as:
\begin{align}
\label{equ:input}
\begin{pmatrix}\mathbf{i}_t
\\ \mathbf{f}_t
\\ \mathbf{o}_t
\\ \mathbf{g}_t
\end{pmatrix}= \begin{pmatrix} \sigma 
\\ \sigma 
\\ \sigma 
\\ \tanh
\end{pmatrix}\mathbf{W}\begin{pmatrix} \mathbf{x}_t
\\ \mathbf{h}_{t-1}
\end{pmatrix}\\ 
\label{equ:cell}
\mathbf{c}_t=\mathbf{f}_t\odot\mathbf{c}_{t-1}+\mathbf{i}_t\odot\mathbf{g}_t\\
\label{equ:att}
\mathbf{a}_t=\mathbf{W}_{a}
\begin{pmatrix} \mathbf{h}_{t-2}
\\ \mathbf{h}_{t-3}
\\ ...
\\ ...
\\ \mathbf{h}_{t-k}
\end{pmatrix}\\
\label{equ:res}
\mathbf{h}_t=\mathbf{o}_t\odot\tanh(\mathbf{c}_t+\mathbf{a}_t)
\end{align}
where $\mathbf{i}_t$, $\mathbf{f}_t$ and $\mathbf{o}_t$ are input, forget and output gate respectively. $\mathbf{c}_t$ is memory cell, $\sigma(\cdot)$ is the sigmoid function. Equations(\ref{equ:input}) - (\ref{equ:cell}) are from original LSTM, $\mathbf{a}_t$ in equation (\ref{equ:att}) is the defined attention gate which summarizes relative contributions in the range from $\mathbf{h}_{t-2}$ to $\mathbf{h}_{t-k-1}$. The hidden state $\mathbf{h}_{t-1}$ is used in original way and attended at each step so that to form a residual (shortcut) connection across timesteps. The attention weights $\mathbf{W}_{a}$ is normalized by $\mathbf{W}_{a}^{(i)}$=$\frac{\mathbf{W}_{a}^{(i)}}{\sum_{j}^{K}\mathbf{W}_{a}^{(j)}}$\footnote{while softmax is more often used here,  we found this is more straightforward and faster in BPTT without losing performance.}. 
In equation (\ref{equ:res}), follow residual network \cite{he2016deep}, element-wise addition is used to form the residual function of attention $\mathbf{a}_t$ which directly effect memory cell $\mathbf{c}_t$ for outputting $\mathbf{h}_{t}$.

By defining an attention gate in RNNs, only $K$ additional differentiable parameters over residual connection are introduced. The optimization can be realized by using standard back-propagation through time (BPTT)\cite{williams1986learning} as regular RNNs. 

\section{Experiments}
\label{sec:experiment}
In this section, we explore the performance of proposed RRA in multiple tasks including the adding problem, pixel-by-pixel MNIST image classification and sentiment analysis on the IMDB dataset. 

Our implementation was based on Theano\footnote{\url{http://www.deeplearning.net/software/theano/}}. We conducted all our experiments on a single Titan Xp with 12G memory. The weights for input-to-hidden layer and hidden-to-output layer were initialized by drawing the uniform distribution $\left [ -\sqrt{\frac{6}{N_{in}+N_{out}}},\sqrt{\frac{6}{N_{in}+N_{out}}} \right ]$ ($N$: number of units). The RNN internal weights $\mathbf{W}$ were orthogonally initialized \cite{saxe2013exact}. The attention weights $\mathbf{W}_a$ were randomly initialized. By default, the attention window size $K$=10, which means the past hidden states from $\mathbf{h}_{t-1}$ to $\mathbf{h}_{t-11}$ are considered at every timestep. Initial learning rate was set to 0.0001 and 0.5 dropout rate was used after recurrent layer. Gradients were clipped to 1 to prevent exploding gradients. All models were configured to have only one recurrent layer and trained with given number of iterations without early stopping. All experimental settings for LSTM and RRA are same. 

\subsection{Adding Problem}
This task was originally defined in \cite{hochreiter1997long} for testing the ability of RNN to capture the long dependencies in a sequential data. The task is asked to add two numbers $x_i$ and $x_j$ that randomly selected from a sequence. For a given sequence with length $S$, each element of this sequence is a pair consisting of two components $(x, m)$, the first one is an actual number $x$ that uniformly sampled at $\mathcal{U}[0,1]$, the second one is an indicator $m$ decides whether to add $x$ (if $m$=$1$) or just ingore $x$ (if $m$=$0$).  There are only two numbers ($x_i$ and $x_j$) in each sequence are marked as 1 for addition: the first number $x_i$ is placed to the first 10\% of sequence, i.e. $i\in[0,\left \lfloor \frac{S}{10} \right \rfloor]$, the second number $x_j$ is from the last 50\% in the sequence, i.e. $j\in [\left \lfloor \frac{S}{2} \right \rfloor, S]$. This leads to a sequence has long-range dependency where only two significant but remote inputs.   A naive strategy is always to predict the target output as 1 regardless of the input sequences \cite{le2015simple,arjovsky2016unitary}, it gives an expected mean squared error (MSE) of 0.167 which is used as baseline to beat. 

\begin{figure}[!htbp]
\centering
\minipage{0.35\textwidth}
  \includegraphics[width=\linewidth]{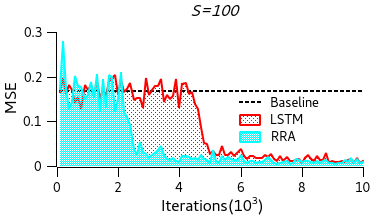}
\endminipage\hfill
\minipage{0.35\textwidth}
  \includegraphics[width=\linewidth]{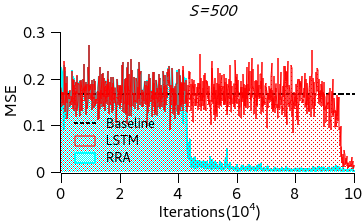}
\endminipage
 \caption{The performance on the adding problem for sequence length $\textit{S}$=$100$ (top) and $\textit{S}$=$500$ (bottom). }
\label{fig:adding}
\end{figure}

We used 128 hidden units for both LSTM and RRA, the batch size was set to 50, the models were optimized with ADADELTA \cite{zeiler2012adadelta}. We generated 100,000 training examples and 10,000 test examples.  Figure \ref{fig:adding} presents the performance of LSTM and RRA on test dataset as we varied sequence length $\textit{S}$.  As we can see,  for $\textit{S}$=$100$, LSTM is able to consistently beat baseline around 4,400 iterations while RRA approximately beats baseline at 2,200 iterations.  As we increased $\textit{S}$ to 500, the task gets harder because the dependency between target output and the two relevant sequence inputs becomes more remote, this requires model is able to capture longer dependencies. In the first 40,000 iterations, both LSTM and RRA struggled to minimize MSE, RRA started to beat baseline after 43,000 iterations, this is significantly faster than LSTM that started to beat baseline after around 92,000 iterations.

Although this task against the advantage of RRA since there are only two significant numbers in each sequence, RRA demonstrates good performance in learning long-range dependencies.
\begin{figure}[!htbp]
\centering
\minipage{0.235\textwidth}
  \includegraphics[width=\linewidth]{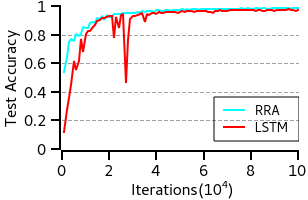}

\endminipage\hfill
\minipage{0.235\textwidth}
  \includegraphics[width=\linewidth]{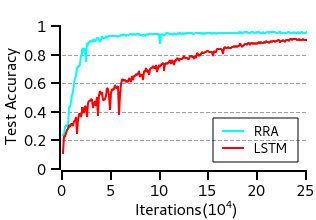}
\endminipage
 \caption{Performance on Pixel-by-Pixel MNIST. Normal MNIST (left) and Permuted MNIST (right).}
\label{fig:mnist}
\end{figure}
\subsection{Pixel-by-Pixel MNIST}
This task is asked to classify MNIST digits \cite{lecun1998gradient} as suggested by \cite{le2015simple}. Each 28-by-28 image in MNIST is treated as sequential data and fed to recurrent network.  This leads to pixel sequences with length of 784.  Two versions of pixel-by-pixel MNIST were considered: (1) normal MNIST that the pixel sequence is read in order from left to right, top to bottom. (2) The pixel sequence is randomly permuted.
We configured both networks to have 256 hidden units, optimizer is replaced with RMSprop which provides more steady improvement on this task for both networks. The training batch size is 50, LSTM is used as baseline to beat as plain RNN has been proved poor performance on such tasks in \cite{le2015simple,arjovsky2016unitary}.
 
Figure \ref{fig:mnist} reports the test accuracy against iterations. On normal pixel-by-pixel MNIST (Figure \ref{fig:mnist}(left)), similar to previous work \cite{arjovsky2016unitary}, both LSTM and RRA show good performance.  RRA achieves 98.58\% that beats LSTM of  97.66\%. Besides, it shows that RRA is able to yield faster convergence, more stable improvement as compared to the standard LSTM. 

The task was configured to be more challenging when we randomly permuted the order of pixels in image. By applying same permutation to each image, the dependencies across pixels become longer than original pixel order. This requires models to learn and remember more complicated dependencies across different timesteps. As shown in Figure \ref{fig:mnist}(right),  RRA shows superior capability in capturing such long and complicated dependencies. It achieves 95.84\% against 91.2\% for LSTM, but again, faster convergence.

We further compared RRA with recent proposed methods: IRNN \cite{le2015simple}, URNN \cite{arjovsky2016unitary} and RWA \cite{ostmeyer2017machine} in Table \ref{tab:mnist_comp}. RRA achieves the state-of-the-art performance on both normal and permuted pixel-by-pixel MNIST. It should be noted that both URNN and RWA are not able to beat LSTM on normal MNIST in their configurations.  Nevertheless, RRA achieves sightly better performance on normal MNIST and outperforms LSTM on permuted MNIST in a certain margin.    
\begin{table}
\centering
\caption{Test accuracy on pixel-by-pixel MNIST}
\label{tab:mnist_comparison}
\begin{tabular}{c|c|c} \hline
Models         & Normal MNIST & Premuted MNIST \\ \hline\hline
IRNN          & 97\%   & 82\%   \\
URNN             & 95.1\%   & 88\%    \\
RWA                    & 98.1\%   &   93.5\%   \\ \hline\hline
LSTM           & 97.66\%       & 91.2\% \\
RRA &98.58\% &95.84\%   \\ \hline 
\end{tabular}
\label{tab:mnist_comp}
\end{table}
\begin{table*}[!htb]
\small
\centering
\caption{Performance comparison on IMDB Review Dataset}
\label{tab:imdb_comparison}
\begin{tabular}{c|c} \hline
Models         & Reported Error Rate \\ \hline\hline
BoW (bnc)\cite{maas-EtAl:2011:ACL-HLT2011} &  12.20\%       \\
BoW(b$\Delta$ t$\acute{c}$) \cite{maas-EtAl:2011:ACL-HLT2011} &11.77\%         \\
LDA \cite{maas-EtAl:2011:ACL-HLT2011}              &   32.58\%     \\
LSA \cite{maas-EtAl:2011:ACL-HLT2011}              &  17.04 \%      \\
Full+BoW \cite{maas-EtAl:2011:ACL-HLT2011}              &   11.67\%     \\
Full+unlabelled+BoW \cite{maas-EtAl:2011:ACL-HLT2011}              &  11.11\%      \\
WRRBM \cite{dahl2012training}     &    12.58\%     \\
WRRBM+BoW(bnc) \cite{dahl2012training}     &    10.77\%    \\
MNB-uni \cite{wang2012baselines} &  16.45\%        \\
MNB-bi \cite{wang2012baselines} &    13.41\%    \\
SVM-uni \cite{wang2012baselines} &     13.05\%   \\
SVM-bi \cite{wang2012baselines} & 10.84\%    \\
NBSVM-uni \cite{wang2012baselines} & 11.71\%    \\
NBSVM-bi \cite{wang2012baselines} & 8.78\%    \\
seq2-bown-CNN\cite{johnson2014effective} & 14.70\%    \\
Paragraph Vector \cite{le2014distributed}& 7.42\%    \\
LSTM with tuning and dropout \cite{dai2015semi}  & 13.50\%    \\
LSTM initialized with word2vec embeddings \cite{dai2015semi} &  10.00\%    \\
LM-LSTM \cite{dai2015semi}  & 7.64\%    \\
SA-LSTM \cite{dai2015semi}  &7.24\%    \\
SA-LSTM with liner gain \cite{dai2015semi} & 9.17\%    \\
SA-LSTM with joint training \cite{dai2015semi} & 14.70\%    \\ 
TS-ATT\cite{yuan2016learning}& 13.75\%    \\ 
SS-ATT\cite{yuan2016learning}& 13.26\%    \\ \hline\hline

LSTM           & 11.63\%        \\
RRA(K=5)     & 11.27\%        \\ 
RRA(K=10) & 11.59\%   \\ 
RRA(K=20) & 12.22\%   \\ 
Bidirectional RRA (K=5) & 9.05\%   \\
\hline 
\end{tabular}
\end{table*}
\subsection{Sentiment Analysis}
To evaluate the performance of RRA on sentiment analysis, we conducted experiments on IMDB review dataset \cite{maas-EtAl:2011:ACL-HLT2011}\footnote{\url{http://ai.stanford.edu/~amaas/data/sentiment/}}. This dataset consists of 100,000 movie reviews from IMDB.  The dataset is split into 75\% for training and 25\% for testing. There are only 25,000 reviews in training reviews are labeled, and the rest of 50,000 are unlabeled, all testing reviews are labeled. In this task,we used the labeled 25,000 training reviews and 25,000 test for binary sentiment classification (positive or negative), thus randomly guessing yields 50\% accuracy. Different to some previous approaches, e.g. Bag-of-Words (BOW) and Latent Dirichlet Allocation (LDA)\cite{blei2003latent} etc., the review sentences are treated as sequential data. This task is particularly challenging because the average review length is 281 and the longest review can reach 2,956 words. This requires our model has strong ability to capture the long-range semantic dependencies among words.  
\vspace{-0.4cm} 
\begin{figure}[!htb]
\centering
\includegraphics[width=0.4\textwidth]{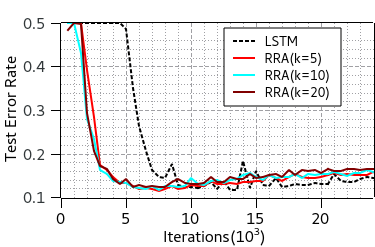}
\caption{Performance on IMDB Review Dataset.}
\label{fig:imdb_test}
\end{figure}

In our experiments, we limited the word vocabulary size to 10,000, all other words were mapped to ``\emph{unk}'' token. We used 128 units for embeddings and 128 units for both LSTM and RRA with ADADELTA\cite{zeiler2012adadelta} optimizer, batch size was set to 16. We tested RRA with different attention window size $K$=$5$, $K$=$10$ and $K$=$20$. Figure \ref{fig:imdb_test} presents the test error against iterations for original LSTM and RRA with different $K$. Each model was trained around 15 epochs without early stopping. We can see that RRA fits the dataset quite well since 4,000 iterations, considerably faster than LSTM. With varied attention window size $K$, we found that the test error is not very sensitive to different $K$, RRA obtains sightly better results when $K$=5. We conjecture that for a certain pattern of sequence (e.g. English sequence in this task), the semantic contributions from previous $K$ hidden states are sufficient to compute the current state. 

In order to compare RRA with recent methods, we add more recently reported baselines. Table \ref{tab:imdb_comparison} shows the performance comparison. It proves that RRA can effectively learn good representations from input word sequence for sentiment classification as compared to previous non-sequential representations, e.g. BoW, LDA and LSA with SVM classifiers. RRA is also highly competitive to recent approaches LM-LSTM and SA-LSTM (which used 1024 units for memory cells, 512 embedding units with 50,000 unlabeled reviews for per-training). It should be noted that our models were solely based on proposed RRA with only 128 hidden units, without using additional unlabeled data for pre-training as well as word2vec embeddings. With bidirectional RRA, the performance of our model is sufficiently close to the state-of-the-art.

\begin{figure}[!htb]
\minipage{0.5\textwidth}
  \includegraphics[width=\linewidth]{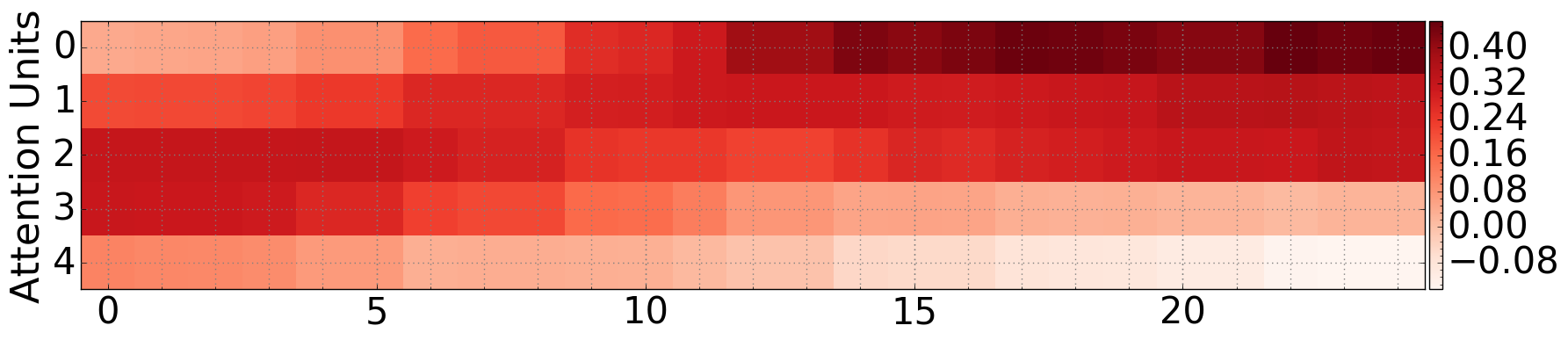}
\endminipage \hfill
\minipage{0.5\textwidth}
  \includegraphics[width=\linewidth]{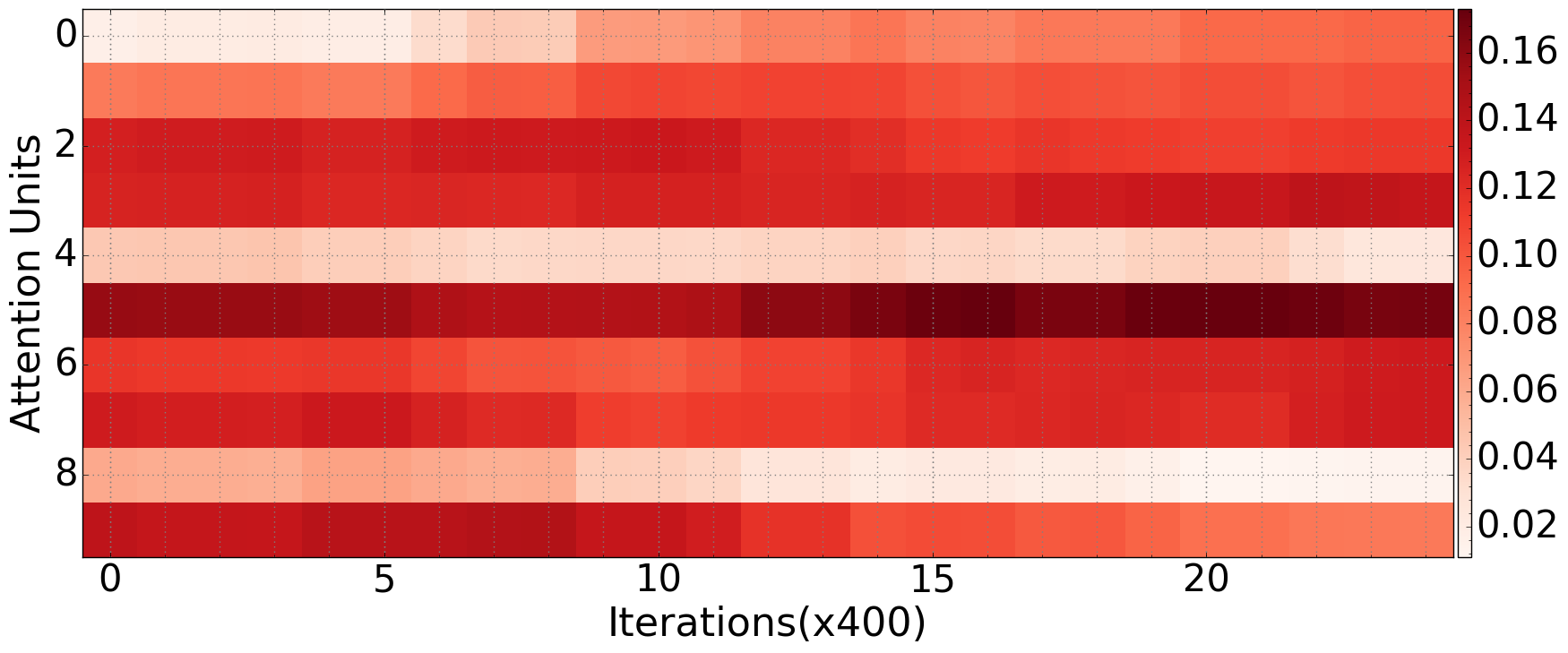}
\endminipage
 \caption{Visualization of normalized attention weights for $K$=5 (top) and $K$=10 (bottom). The attention unit index 0 corrsponding to $\mathbf{W}_a^{(1)}$, the weight that is assigned to $\mathbf{h}_{t-2}$.}
\label{fig:vis}
\end{figure}
We also visualized the attention weights in the case of $K$=5 and $K$=10 respectively in Figure \ref{fig:vis}. The evolution of normalized weights of attention units suggests that attention gate learns to control the relative contributions from previous hidden states from $\mathbf{h}_{t-2}$ to $\mathbf{h}_{t-K-1}$. They are explicitly considered in predicting $\mathbf{h}_t$, this is contrast with standard RNN/LSTM and other variants where history information indirectly considered via $\mathbf{h}_{t-1}$.

\subsection{Discussion}
\noindent  \textbf{RRA alleviates gradient vanishing}
In BPTT,  gradient vanishing when gradient $\frac{\partial L}{\partial \mathbf{W}}$=$\sum\frac{\partial L}{\partial z}\frac{\partial z}{\partial \mathbf{h}_T}\frac{\partial \mathbf{h}_T}{\partial \mathbf{h}_{T-1}}\cdots \frac{\partial \mathbf{h}_0}{\partial \mathbf{W}}$ is close 0. Because it sums each gradient contribution from every timestep, the dependency across timesteps cannot be captured if the gradient contribution is 0. RRA explicitly enforces short-cut connection across timestep and directly passes error signal through $\mathbf{h}_T$ to $\mathbf{h}_{T-K}$. The attention over residual connection enables to control the relative contribution across multiple timesteps to alleviate gradient become to 0, particularly in learning dependencies from long and complex sequence. Our experiments in Figure \ref{fig:adding}, \ref{fig:mnist} and \ref{fig:imdb_test} have  demonstrated the stability of RRA in learning long and complex sequence.     

\noindent  \textbf{Relation to related work}
There are some RNN variants have been recently proposed to address gradient vanishing problem in recurrent networks. IRNN \cite{le2015simple} is an RNN that is composed of ReLUs and initialized with an identity weight matrix, URNN \cite{arjovsky2016unitary} uses a unitary hidden-to-hidden matrix by generalizing the orthogonal matrices to the complex domain. Differently, this work focuses on explicitly use multiple previous hidden states via residual connection with attention. 
Higher order RNN (HORNNs)\cite{soltani2016higher} is proposed for language modeling which is similar to our work but the key differences are existed: (1) RRA uses $\mathbf{h}_{t-1}$ as regular RNN so that to form a residual connection with attention while HORNN directly considers $\mathbf{h}_{t-1}$ to $\mathbf{h}_{h-K}$. (2) RRA introduces much less parameters, e.g., when each unit is required to consider the past 3 states,  RRA only introduces 2 additional parameters while HORNN introduces 0.3 millions more weights compared to a plain RNN. 
Recurrent Weighted Average (RWA)\cite{ostmeyer2017machine} also explores attention in RNN. But the difference is that RWA performs a weighted average over $\mathbf{h}_{1}$ to $\mathbf{h}_{t-1}$ when computing each $\mathbf{h}_t$. RRA is more flexible by considering $K$+$1$ past states with residual attention. 

\noindent  \textbf{Limitation of RRA}
Although RRA shows its ability in capturing long-range dependencies across timesteps with faster convergence, more stable training compared to a standard LSTM on multiple tasks, it also has limitation: training speed is sightly slower than standard LSTMs, e.g., on permuted MNIST, LSTM took average 394s for one epoch while RRA($K$=5) took 760s and RRA($K$=10) took 773.6s.  We conjecture that additional time is spent to compute the derivative of residual attention, and pass the error signal from current states to the states that are several step far apart directly. However, it should be noted that, all our experiments did not use early stopping, when it is applied to RRA and LSTM, RRA can finish the training and stop much earlier than LSTM. 
\section{Conclusion}
\label{sec:conclusion}
In this paper we introduced RRA to learn long-term dependencies from sequential data. The residual shortcut connection can effectively pass error signal across timesteps that are several apart away so that to prevent gradient vanishing problem. The defined attention mechanism over timesteps provides a more natural way to summarize the individual contribution of the past hidden states in predicting future hidden states. We compared RRA to a standard implementation of LSTM. RRA shows superior performance, more stable training and fast convergence on the adding problem, pixel-by-pixel MNIST classification and sentiment analysis.  Although without using additional mechanism, e.g. word2vec embedding, pre-training with unlabeled data, RRA demonstrates competitive performance as compared to recent methods. Future work will extend RRA on different sequence learning scenarios including machine translation, speech recognition etc.. 
\section{Acknowledgements}
We thank Mathias Niepert and Brandon Malone for their discussions and suggestions on this work.

\bibliographystyle{aaai}
\bibliography{references}

\end{document}